\documentclass[lettersize,journal]{IEEEtran}
\usepackage{amsmath,amsfonts}
\usepackage{algorithmic}
\usepackage{algorithm}
\usepackage{array}
\usepackage[caption=false,font=normalsize,labelfont=sf,textfont=sf]{subfig}
\usepackage{textcomp}
\usepackage{stfloats}
\usepackage{url}
\usepackage{verbatim}
\usepackage{graphicx}
\usepackage{cite}
\usepackage[pagebackref=false,breaklinks=true,colorlinks,bookmarks=false]{hyperref}
\usepackage{array,tabularx}
\usepackage{booktabs}
\usepackage{multirow}
\usepackage{makecell}
\usepackage{bm}
\begin{document}

\title{Transmission Line Defect Detection Based on UAV Patrol Images and Vision-language Pretraining}

\author{Ke Zhang,~\IEEEmembership{Member,~IEEE,}
        Zhaoye Zheng,
        Yurong Guo,
        Jiacun Wang,~\IEEEmembership{Senior Member,~IEEE,}
        Jiyuan Yang,
        and Yangjie Xiao

\thanks{Research supported by the National Natural Science Foundation of China (NSFC) under grant numbers 62076093, 61871182, 62206095, by the Fundamental Research Funds for the Central Universities under grant numbers 2022MS078, 2023JG002, 2023JC006.}

\thanks{K. Zhang, Z. Zheng, Y. Guo, J. Yang and Y. Xiao are with the Department of Electronic and Communication Engineering, North China Electric Power University, Hebei, 071003, P. R. China. K. Zhang is also with Hebei Key Laboratory of Power Internet of Things Technology, North China Electric Power University, Baoding 071003, Hebei, China.

J. Wang is with the Computer Science and Software Engineering Department, Monmouth University, West Long Branch 07764, USA. J. Wang is the corresponding author (e-mail: jwang@monmouth.edu).}}

\markboth{Journal of \LaTeX\ Class Files,~Vol.~14, No.~8, August~2021}%
{Shell \MakeLowercase{\textit{et al.}}: A Sample Article Using IEEEtran.cls for IEEE Journals}


\maketitle

\begin{abstract}
Unmanned aerial vehicle (UAV) patrol inspection has emerged as a predominant approach in transmission line monitoring owing to its cost-effectiveness. Detecting defects in transmission lines is a critical task during UAV patrol inspection. However, due to imaging distance and shooting angles, UAV patrol images often suffer from insufficient defect-related visual information, which has an adverse effect on detection accuracy. In this article, we propose a novel method for detecting defects in UAV patrol images, which is based on vision-language pretraining for transmission line (VLP-TL) and a progressive transfer strategy (PTS). Specifically, VLP-TL contains two novel pretraining tasks tailored for the transmission line scenario, aimimg at pretraining an image encoder with abundant knowledge acquired from both visual and linguistic information. Transferring the pretrained image encoder to the defect detector as its backbone can effectively alleviate the insufficient visual information problem. In addition, the PTS further improves transfer performance by progressively bridging the gap between pretraining and downstream defection detection. Experimental results demonstrate that the proposed method significantly improves defect detection accuracy by jointly utilizing multimodal information, overcoming the limitations of insufficient defect-related visual information provided by UAV patrol images.

\end{abstract}

\begin{IEEEkeywords}
Power transmission line, defect detection, vision-language pretraining, transfer learning.
\end{IEEEkeywords}

\section{Introduction}
\IEEEPARstart{W}{ith} the rapid development of power system construction, transmission line lengths are expanding at an unprecedented rate. The conventional manual inspection mode no longer meets the growing demands for transmission line inspections. In such context, unmanned aerial vehicle (UAV) patrol inspection, supported by advances in UAV control technologies, has been widely adopted as a new predominant mode\cite{ref1, ref36, ref37}. One of the core components in a UAV patrol inspection system is a deep learning-based defect detector, which is used to locate and identify defects captured in the patrol images\cite{ref2, ref3}. Therefore, improving the detection accuracy of defect detectors is crucial for enhancing the performance of UAV patrol inspection systems and ensuring the stable operation of transmission lines.

However, the performance of existing detectors trained with UAV patrol images remains suboptimal. Specifically, due to the imaging distance and shooting angles of UAV patrol images, components and defects in the images generally exhibit characteristics such as small size, partial occlusion, and complex backgrounds\cite{ref4, ref5, ref6}. Consequently, these images fail to provide sufficient defect-related visual information, causing the detector trained on such data cannot acquire essential knowledge for accurate detection. Therefore, an intuitive solution is to jointly utilize multimodal information to supplement the insufficiency of visual information. 

\begin{figure}[!t]
\centering
\includegraphics[width=3.4in]{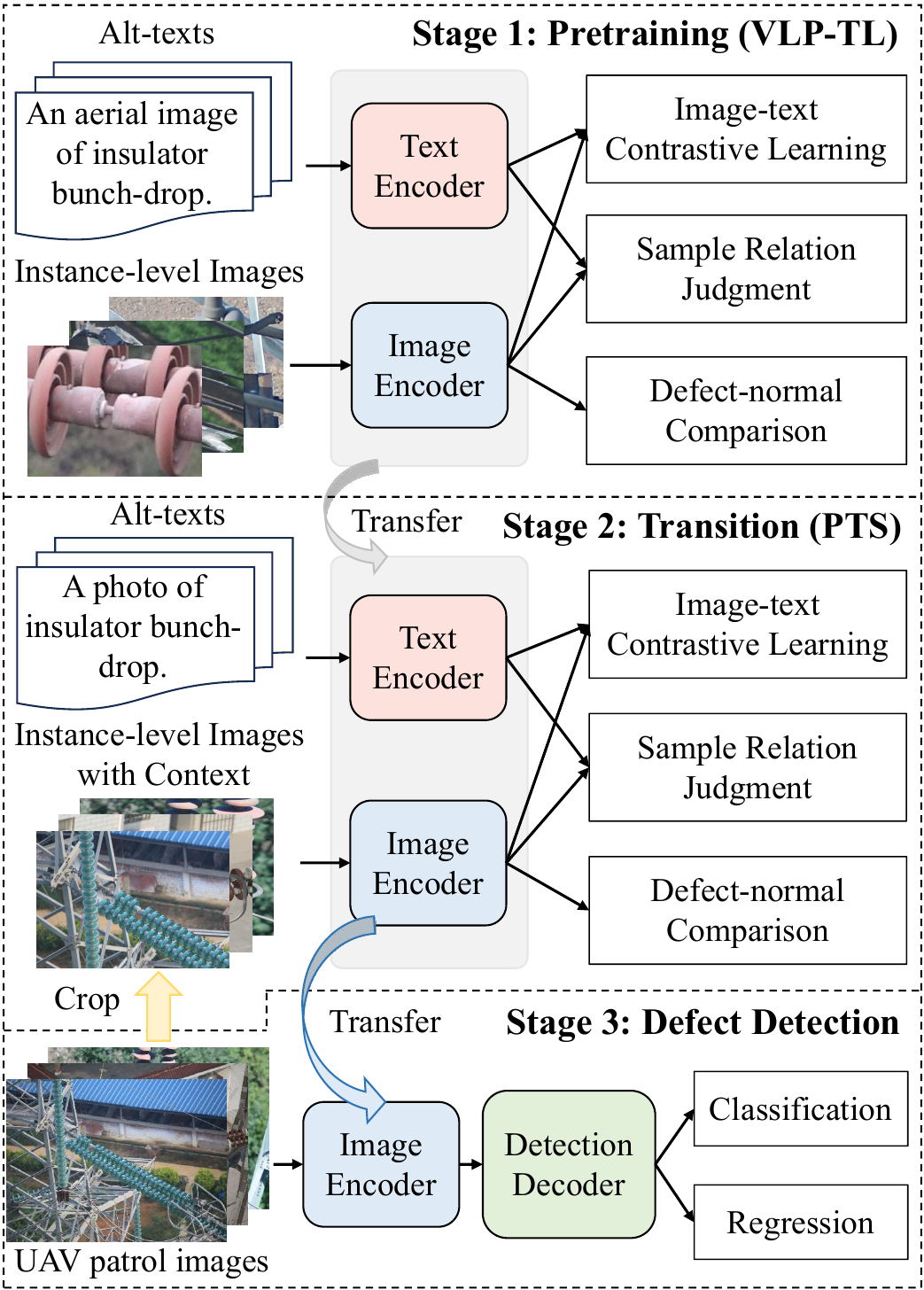}
\caption{An overview of the proposed method.}
\label{fig_1}
\end{figure}

In the last few years, vision-language pretraining (VLP) has achieved impressive few-shot and even zero-shot recognition performance by learning transferable feature representations from web-scale image-text pairs\cite{ref15, ref16}. The success of VLP comes from harnessing the complementary nature of the two different modalities (image and text) to provide additional information that enables the model to learn more comprehensive knowledge\cite{ref35,ref53}. Inspired by this, we propose a transmission line defect detection method from UAV patrol images based on VLP. The overview of the proposed method is shown in Fig.~\ref{fig_1}. Firstly, we design VLP for transmission lines (VLP-TL) to pretrain an image encoder capable of extracting features tailored for the transmission line scenario by learning from the rich visual and linguistic information contained in transmission line image-text pairs. Through the VLP-TL, the image encoder acquires extensive professional knowledge, thus, the problem of insufficient defect-related visual information provided by UAV patrol images is significantly mitigated. Subsequently, we propose a progressive transfer strategy (PTS) to further improve the performance when transferring the pretrained image encoder to defect detection. 

The main contributions of this article are as follows. 
\begin{enumerate}
\item{We present the VLP-TL algorithm, which incorporates two pretraining tasks specifically designed for transmission lines. Then, we build a multimodal dataset composed of transmission line image-text pairs to implement VLP-TL. By fully exploiting visual and linguistic information within the dataset, VLP-TL can pretrain an image encoder with abundant knowledge as the backbone of the defect detector.}
\item{We develop PTS, a transfer learning strategy, to mitigate performance degradation caused by the inconsistency between pretraining and downstream defect detection by adding an intermediate transition stage to achieve a smoother transition between pretraining and defect detection.}
\item{After the pertaining and transition stage, we build a defect detector based on the image encoder and train it with the transmission line defect detection dataset (TLDD dataset). The experimental results show that our detector can effectively improve the defect detection accuracy in UAV patrol images.}
\end{enumerate}

The rest of this article is organized as follows. Section~\ref{sec2} introduces some related works. Section~\ref{sec3} describes the proposed method. Section~\ref{sec4} demonstrates the experimental results to verify the effectiveness of our method. Finally, Section~\ref{sec5} concludes this work and discusses the future work. 
 
\section{Related Works} \label{sec2}
Although the images collected in UAV inspections have high resolution, the components and defects within them often occupy only very small areas, accompanied by partial occlusion and complex backgrounds. Such images provide limited defect-related visual information, making it difficult for detectors trained on them to learn sufficient knowledge to support accurate defect detection. To address this problem, this article leverages linguistic information through VLP to supplement the insufficient visual information. Furthermore, a transfer learning strategy is proposed to bridge VLP and defect detection. This section introduces previous works related to this article.

\subsection{Vision-language Pretraining}
VLP is a type of pretraining paradigm that has prevailed in recent years\cite{ref17}. It uses semantic-rich texts instead of predefined labels to provide supervision signals for pretraining an image encoder with abundant knowledge. The pretrained image encoder can subsequently be transferred as the backbone of downstream task models, achieving amazing performance even with limited training samples\cite{ref29, ref30, ref31}. According to the application scenarios, existing VLP algorithms can be divided into two types: 1) general-purpose VLP algorithms and 2) domain-specific VLP algorithms.

General-purpose VLP algorithms are typically based on pretraining tasks such as image-text contrastive learning and leverage large-scale image-text pairs collected from the internet for pretraining. After pertaining, a wide range of public datasets covering a wide range of downstream tasks are used to evaluate the effect of pretraining. These algorithms strive to achieve consistent improvements across various downstream tasks by scaling up model size\cite{ref46}, improving data quality\cite{ref47}, and refining pretraining tasks\cite{ref21}. However, for application scenarios of special domains, directly transferring image encoders pretrained by general-purpose VLP algorithms does not yield satisfactory results due to the lack of consideration for data privacy and uniqueness. Therefore, another line of works focuses on designing domain-specific VLP algorithms for specialized domains such as remote sensing\cite{ref34}, biomedicine\cite{ref32}, industrial inspection\cite{ref18}, etc. They customize dedicated VLP algorithms according to the characteristics or requirements of a certain domain to learn professional knowledge from domain-specific image-text pairs\cite{ref33}. And these VLP algorithms are evaluated only on a few specified downstream tasks of interest to the domain. Unfortunately, there are no domain-specific VLP algorithms for the domain of electric power, particularly for the transmission line scenario.

Therefore, to pretrain an image encoder suitable for processing UAV patrol images in this scenario, we propose the VLP-TL algorithm. Specifically, we design two transmission line-specific pretraining tasks based on the implicit relations between samples to efficiently utilize the visual and linguistic information, thereby acquiring richer knowledge. Benefiting from information from both modalities, the pretrained image encoder is then transferred as the backbone of the detector, which can effectively overcome the problem of insufficient defect-related visual information in UAV patrol images and extract more expressive features for defect detection.

\subsection{VLP-based Transfer Learning}
Transfer learning aims to improve the performance of a downstream model by transferring the knowledge acquired during pretraining. For instance, in defect detection research of transmission lines,  a common transfer learning approach is to use an image encoder pretrained on ImageNet\cite{ref43} as the backbone of the defect detector to improve training effectiveness and efficiency\cite{ref12,ref13,ref14}. With the advancement of VLP, researchers have realized that simply using a VLP-based image encoder as the backbone for downstream tasks cannot yield optimal transfer performance due to differences in data distribution and training objectives\cite{ref48}. To better transfer the rich knowledge learned by VLP, VLP-based transfer learning strategies have also become a research hotspot. Based on whether new parameters are introduced beyond the pretrained backbone and task-specific decoder (e.g., neck and head in detectors) in the downstream model, VLP-based transfer learning strategies can be divided into two types: 1) learnable parameter-based and 2) training skill-based.

learnable parameter-based strategies train a small number of extra learnable parameters to capture downstream task-related knowledge. The introduction of these parameters mainly takes the form of prompt vectors\cite{ref29,ref50} and adapter blocks\cite{ref49,ref51}. On the other hand, training skill-based strategies do not introduce additional parameters but fine-tune the entire pretrained image encoder or task-specific decoder on downstream datasets with elaborate training skills. For example, Lin et al.\cite{ref53} introduced category names in the language modality as additional training samples to the training of downstream classification tasks, effectively improving few-shot classification performance. Goyal et al.\cite{ref52} proposed finetuning like you pretrain (FLYP), which replaces the training objective of downstream classification tasks with the pretraining objective of contrastive learning to achieve higher classification accuracy. However, regardless of the type of transfer learning strategy, their targeted downstream tasks are mainly image classification. Unlike detection tasks, classification tasks use instance-level images (where the sample occupies most of the image area) and even not require task-specific decoders. As a result, these existing strategies cannot be directly applied to downstream tasks such as defect detection.

To enhance the transfer performance of the pretrained image encoder in defect detection, this article introduces an intermediate transition stage between pretraining and defect detection. By maintaining the pretraining tasks unchanged while further optimizing the image encoder with image styles closer to defect detection, the image encoder is progressively transferred to defect detection.

\subsection{Transmission Line Defect Detection}
Common transmission line defects involve component damage as well as external interference that poses some hidden dangers. This article focuses on larger-scale defects in transmission lines, such as fitting damage, insulator damage, and foreign bodies. To detect these defects, existing methods improve general-purpose object detectors that perform well on public datasets according to the data distribution or visual characteristics of defects, and then use private defect detection datasets to train dedicated defect detectors\cite{ref6,ref11,ref26}. Among these previous works, some improvements take the form of introducing more sophisticated network structures to help the detector better capture the professional knowledge hidden in UAV patrol images. For example, Bao et al.\cite{ref22} equipped the defect detector with an end-to-end coordinate attention module, enabling the detector to focus on key components rather than complex backgrounds. Guo et al.\cite{ref23} designed a new feature enhancement structure to obtain a more robust feature representation for detecting corrosion defects in fittings. Meanwhile, some improvements aim to help the model better learn professional knowledge by improving essential factors in training, such as training data and training pipeline. For instance, Li et al.\cite{ref25} introduced an intersection‐over‐union‐based sampling strategy to alleviate the problem of class imbalance. Hao et al.\cite{ref7} adopted a three-step training pipeline to transfer the prior knowledge about insulator parallel-gap faults into a defect detector. However, these works only utilize the limited defect-related visual information in UAV patrol images to train detectors, thus, the knowledge that detectors can learn is constrained. Realizing this problem, some researchers have turned to other forms of supplementary information. Choi et al.\cite{ref9} presented a multimodal image feature fusion module that introduces visual information provided by infrared images while utilizing conventional visible-light images collected by UAVs. Falahatnejad et al.\cite{ref63} resorted to a super-resolution-based approach to help the detector access more subtle information in images. Zhang et al.\cite{ref28} employed self-supervised learning to leverage the visual information contained in a large number of unlabeled normal components, injecting the latent knowledge into the backbone of the detector. Nevertheless, these methods still cannot completely get rid of the problem of insufficient defect-related visual information. Specifically, whether defects in infrared images and images obtained through super-resolution, or unlabeled normal components in visible-light images, they still exhibit visual characteristics such as small size, partial occlusion, or complex background.

In this article, we endeavor to solve this problem by introducing linguistic information. Intuitively, by aligning an image of shockproof hammer intersection with the textual description "two shockproof hammers closely adjacent", the knowledge about this defect can be learned more clearly, even when the shockproof hammers in the image are small or occluded. Through VLP, we successfully utilize linguistic information to train a detector's backbone, which improves defect detection accuracy without compromising the inference speed and detector size.

\section{Method} \label{sec3}
In this section, we introduce our method in detail. It employs a three-stage training pipeline, as shown in Fig.~\ref{fig_1}: 1) pretraining; 2) transition; and 3) defect detection. First, a transmission line-specific VLP algorithm, namely VLP-TL, is used to pretrain a dedicated image encoder by leveraging both visual and linguistic information. Second, a transfer learning strategy called PTS is implemented to facilitate a smooth transition between pretraining and downstream defect detection. Finally, the obtained image encoder is transferred to defect detection by serving as the backbone of the defect detector.

\subsection{Pretraining with VLP-TL}
We propose the VLP-TL algorithm to efficiently utilize transmission line image-text pairs and inject the contained knowledge into the detector's backbone, as shown in Stage 1 of Fig.~\ref{fig_1}.

\emph{1) Pretraining dataset curation:} In transmission line inspection, inspectors typically provide a professional description in natural language for each defect-containing image to ensure accurate defect annotation. However, such a description typically only characterizes a small defect region (often covering less than 10\% of the image area). These image-text pairs cannot be directly applied to VLP, as the extensive background areas would severely interfere with the alignment between the two modalities.

Therefore, we reorganize the data through three steps to construct a multimodal dataset suitable for VLP. 
First, we annotate and crop the original UAV patrol images by category to obtain numerous instance-level images of defect and normal samples. Additionally, we reserve the category annotation for each image for the convenience of implementing pretraining tasks.
Second, we fill predefined templates with category annotations while manually refining the description provided by inspectors to generate diverse alt-texts for each category, thereby creating an alt-text pool.
Finally, we produce image-text pairs by randomly assigning an alt-text from the alt-text pool to each instance-level image according to its category.
Through the above process, we construct a multimodal dataset where each image-text pair describes a sample.

\emph{2) VLP-TL algorithm:} To enhance pretraining efficiency, VLP-TL employs a two-tower model architecture (\emph{i.e.}, a model composed of an image encoder and a text encoder) initialized with existing open-source pretrained weights\cite{ref39}. On this basis, two novel pretraining tasks designed for transmission line samples are introduced to mine the potential knowledge in the multimodal dataset. 

\begin{figure}[!t]
\centering
\includegraphics[width=2.7in]{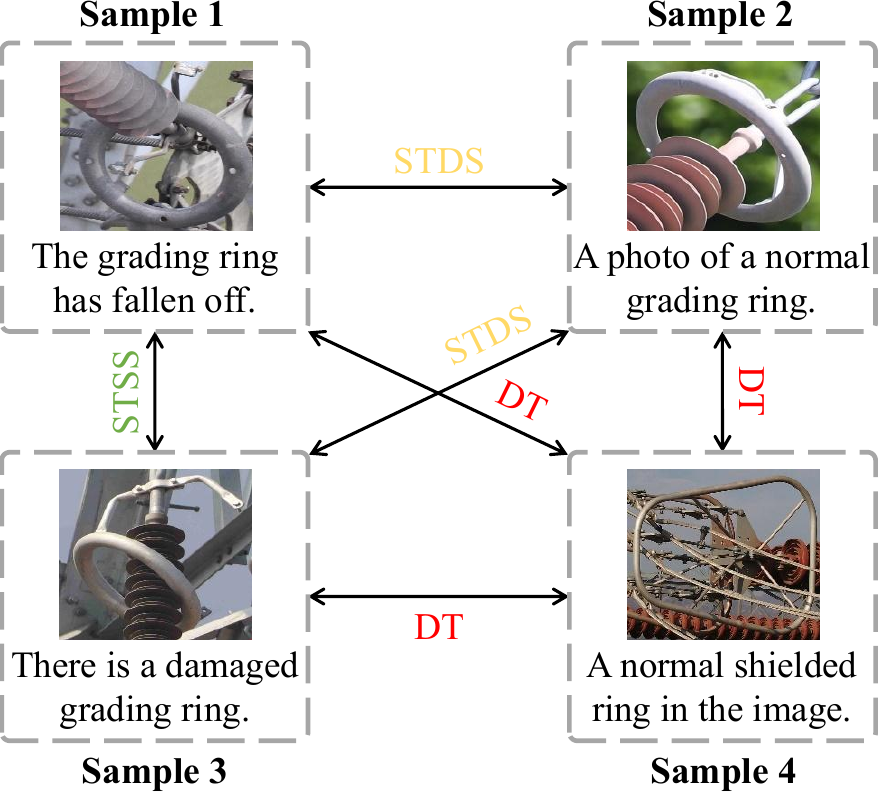}
\caption{Relations between different transmission line samples. STSS, STDS, and DT are abbreviations for same type and same status, same type but different status, and different type, respectively. }
\label{fig_2}
\end{figure}
\begin{figure}[!t]
\centering
\includegraphics[width=3.3in]{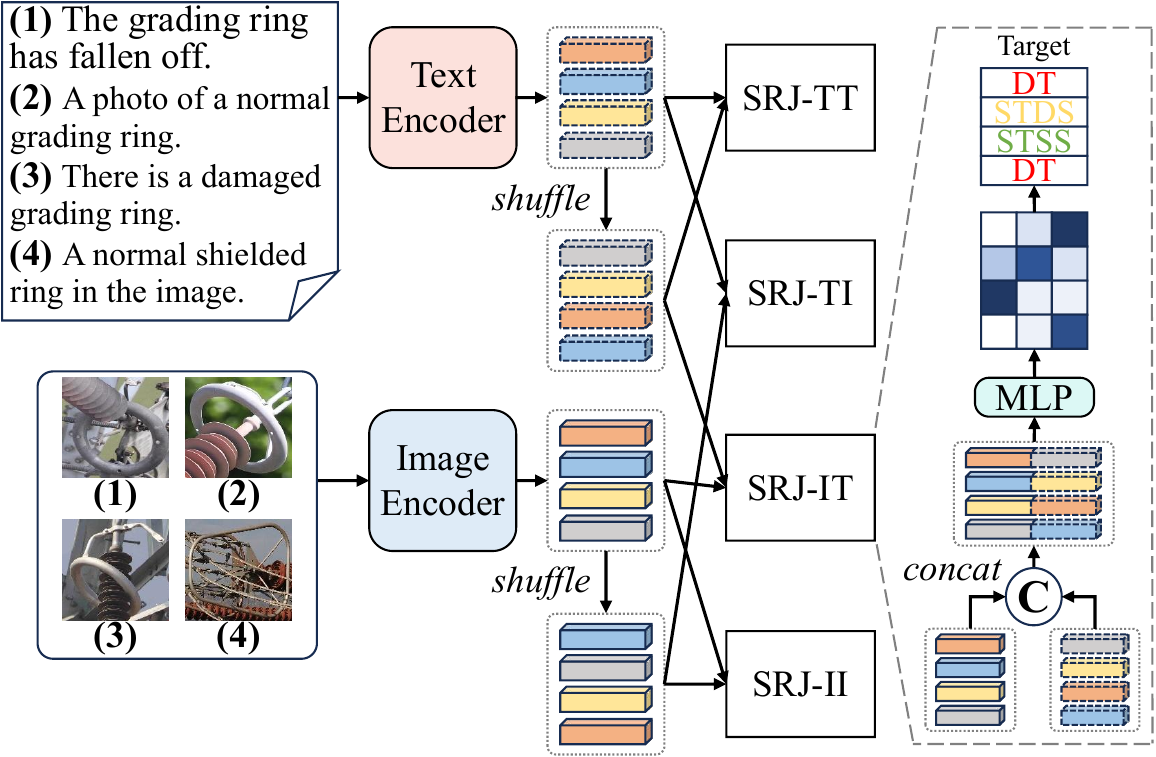}
\caption{The implementation of SRJ.}
\label{fig_3}
\end{figure}

\textbf{Sample relation judgment (SRJ).}
According to the categories of two samples, different samples in public datasets such as ImageNet\cite{ref43} typically exhibit only two relations, \emph{i.e.}, same and different. In contrast, in the transmission line scenario, there are three probable relations between two samples, as shown in Fig.~\ref{fig_2}. a) Same type and same status (STSS): The two samples belong to the same category, in other words, they have the same component type and the same status. For example, the relation between Sample 1 and Sample 3 is STSS, because both of them are grading rings and exhibit defect status. b) Same type but different status (STDS): Although the two samples belong to different categories, they share the same component type. For example, the relation between Sample 1 and Sample 2 is STDS, because both samples are grading rings, but one is defect while the other is normal. c) Different type (DT): Two samples belong to different categories and have distinct component types. For example, the relation between Sample 1 and Sample 4 is DT, because Sample 1 is a grading ring while Sample 4 is a shielded ring. Specially, for external interference defects (\emph{e.g.}, bird nests and foreign bodies), only STSS and DT relations may exist between them and other samples. Learning to classify the relation between two samples into the above three cases facilitates the discovery of cross-component distinctions while capturing both shared features and subtle variations among samples of the same component type. The knowledge acquired in this way can reduce both misjudgments on component types and false alarms on normal components during defect inspection. Inspired by this, we propose the SRJ pretraining task to learn how to judge the relation between two arbitrary samples, as shown in Fig.~\ref{fig_3}.

Given a batch of samples, we denote the images in the batch as $\bm{I}=[\bm{i}_1,\bm{i}_2,\cdots,\bm{i}_n]$ and their alt-texts as $\bm{T}=[\bm{t}_1,\bm{t}_2,\cdots,\bm{t}_n]$, where $n$ is the batch size. We feed the images into the image encoder and the alt-texts into the text encoder to extract features from both modalities:
\begin{equation}
\bm{V}=f_{i}(\bm{I}) \quad and \quad \bm{L}=f_{t}(\bm{T}) \, ,
\end{equation}
where $f_{i}\left ( \cdot  \right )$ and $f_{t}\left ( \cdot  \right )$ represent the image encoder and the text encoder, respectively. $\bm{V}=\left [\bm{v}_1, \bm{v}_2, \cdots, \bm{v}_n \right]$ is the image feature matrix, where $\bm{v}_i$ represents the image feature vector of the $i$-th image in $\bm{I}$. $\bm{L}=\left [ \bm{l}_1, \bm{l}_2, \cdots, \bm{l}_n  \right ]$ is the text feature matrix, where $\bm{l}_i$ represents the text feature vector of the $i$-th alt-text in $\bm{T}$. 

Then, shuffle feature vectors within $\bm{V}$ and $\bm{L}$ to obtain two shuffled feature matrices: 
\begin{equation}
\bm{V}^{\prime}=\text{Shuffle}(\bm{V}) \quad and \quad \bm{L}^{\prime}=\text{Shuffle}(\bm{L}) \, ,
\end{equation}
where $\text{Shuffle}\left ( \cdot  \right )$ represents the shuffle operation. $\bm{V}^{\prime}$ and $\bm{L}^{\prime}$ are the shuffled image feature matrix and the shuffled text feature matrix, respectively. 

Next, select one original feature matrix and one shuffled feature matrix to judge the relations between two samples corresponding to feature vectors at the same positions. According to different modality combinations, SRJ can be divided into four parallel subtasks: a) SRJ-IT: implemented with $\bm{V}$ and $\bm{L}^{\prime}$; b) SRJ-TI: implemented with $\bm{L}$ and $\bm{V}^{\prime}$; c) SRJ-II: implemented with $\bm{V}$ and $\bm{V}^{\prime}$; d) SRJ-TT: implemented with $\bm{L}$ and $\bm{L}^{\prime}$. Taking SRJ-IT as an example, we concatenate $\bm{V}$ and $\bm{L}^{\prime}$ and use a multi-layer perceptron (MLP) shared by the four subtasks to predict the relations:
\begin{equation}
\bm{P}_\text{srj-it} = \text{MLP}\left(\text{Concat}\left( \bm{V}, \bm{L^{\prime}} \right) \right),
\end{equation}
where $\text{MLP}\left ( \cdot  \right )$ represents a MLP module, and $\text{Concat}\left ( \cdot  \right )$ represents the concatenation operation. $\bm{P}_\text{srj-it} \in \mathbb{R}^{n \times 3}$ is the prediction matrix.

Subsequently, the loss of SRJ-IT can be calculated as:
\begin{equation}
L_\text{srj-it} = \text{CE}\left(\bm{P}_\text{srj-it}, \bm{Y}_\text{srj-it}\right),
\end{equation}
where $\text{CE}\left ( \cdot  \right )$ represents the cross entropy loss function. $\bm{Y}_\text{srj-it}$ is the target matrix of SRJ-IT obtained by pre-determining the relations between two samples corresponding to feature vectors at the same positions in $\bm{V}$ and $\bm{L}^{\prime}$ according to their category annotations which are reserved during the construction of the multimodal dataset.

Finally, we combine the loss of four subtasks to calculate the loss of SRJ:
\begin{equation}
L_\text{srj} =\left( L_\text{srj-it} + L_\text{srj-ti} + L_\text{srj-ii} + L_\text{srj-tt} \right) /4 \, .
\end{equation}
where $L_\text{srj-ti}$, $L_\text{srj-ii}$ and $L_\text{srj-tt}$ are the loss of SRJ-TI, SRJ-II and SRJ-TT, respectively.

\begin{figure}[!t]
\centering
\includegraphics[width=3.4in]{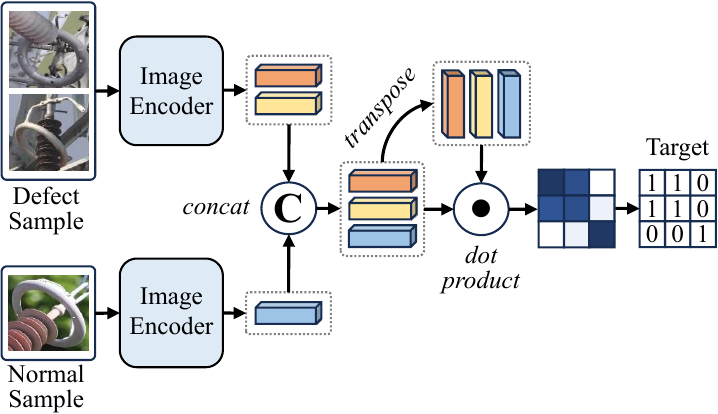}
\caption{The implementation of DNC, taking the case of grading rings as an example.}
\label{fig_4}
\end{figure}

\textbf{Defect-normal comparison (DNC).} We further propose the DNC pretraining task, which requires the image encoder to compare multiple samples with the same component type and determine whether any two samples share identical states, as shown in Fig.~\ref{fig_4}. As can be found that each component type is associated with distinct categories. For instance, the "grading ring" component type is associated with categories such as "normal grading ring" and "grading ring damage". Samples with the same component type and from different categories share substantial visual commonalities, while their differences primarily stem from variations in status, background, and viewing angles. As can be seen from the images of Sample 1 to Sample 3 in Fig.~\ref{fig_2}, they all depict a cyclic structure at the end of an insulator string. Beyond background and viewing angle differences, the key distinction lies in their status: Sample 1 and Sample 3 show detached cyclic structures, while Sample 2 displays a stably fixed structure surrounding the insulator string. Therefore, comparing samples with the same component type reveals discriminative visual cues of different status, which enables more accurate defect detection and reduces false alarms.

For a given batch of samples, we first filter images of defect and normal samples belonging to a specific component type using the reserved category annotations. We denote the images of filtered defect samples as $\bm{I}^\text{d}=[\bm{i}^\text{d}_1,\bm{i}^\text{d}_2,\cdots,\bm{i}^\text{d}_k]$ and the images of filtered normal samples as $\bm{I}^\text{n}=[\bm{i}^\text{n}_1,\bm{i}^\text{n}_2,\cdots,\bm{i}^\text{n}_q]$, where $k$ and $q$ are the number of defect and normal samples. Then we feed $\bm{I}^\text{d}$ and $\bm{I}^\text{n}$ into the image encoder to extract features of defect and normal status:
\begin{equation}
\bm{V}^\text{d}=f_{i}(\bm{I}^\text{d}) \quad and \quad \bm{V}^\text{n}=f_{i}(\bm{I}^\text{n}) \, ,
\end{equation}
where $\bm{V}^\text{d}=[\bm{v}^\text{d}_1,\bm{v}^\text{d}_2,\cdots,\bm{v}^\text{d}_k]$ and $\bm{V}^\text{n}=[\bm{v}^\text{n}_1,\bm{v}^\text{n}_2,\cdots,\bm{v}^\text{n}_q]$ represent the image feature matrix of defect and normal samples, and each feature vectors within $\bm{V}^\text{d}$ and $\bm{V}^\text{n}$ corresponding to the images at the same position of $\bm{I}^\text{d}$ and $\bm{I}^\text{n}$.

Next, we concatenate $\bm{V}^\text{d}$ and $\bm{V}^\text{n}$ to obtain a joint image feature matrix encompassing both defect and normal status:
\begin{equation}
\bm{V^\text{j}} = \text{Concat}\left( \bm{V}^\text{d}, \bm{V}^\text{n} \right)\, ,
\end{equation}
where $\bm{V^\text{j}}$ is the joint image feature matrix. 

After that, we compute the similarity between every two feature vectors within $\bm{V^\text{j}}$, which can be formulated as:
\begin{equation}
\bm{S} = \bm{V^\text{j}}\odot(\bm{V^\text{j}})^{\top}= (s_{ij})_{(k+q)\times(k+q)}\, ,
\end{equation}
where $\odot$ represents dot product, and $\bm{S}$ is the similarity matrix. $s_{ij}$ is the element at the $i$-th row and the $j$-th column of $\bm{S}$, it represents the similarity score between the $i$-th and $j$-th feature vectors in $\bm{V^\text{j}}$.

Then, the loss of DNC on the specific component type can be calculated as:
\begin{equation}
L_\text{dnc}^c = \text{BCE}\left(\bm{S}, \bm{Z}\right)\, ,
\end{equation}
where $\text{BCE}\left ( \cdot  \right )$ represents the binary cross entropy loss function. $c$ represents the index of the currently used component type. $\bm{Z}=(z_{ij})_{(k+q)\times(k+q)}$ is the target matrix, whose element $z_{ij}$ is given as follows:
\begin{equation}
z_{ij} = 
\begin{cases}
1 & \text{$1\leq i,j \leq k$ or $k< i,j \leq k+q$} \\
0 & \text{otherwise} \, 
\end{cases}\, ,
\end{equation}

In practice, DNC operates across multiple component types, comparing defect and normal samples for each component type in parallel. The total loss of DNC is computed as the weighted sum of losses across all component types:
\begin{equation}
L_\text{dnc} =\sum_{c=1}^{r}\alpha_c L_\text{dnc}^c \, .
\end{equation}
where $r$ is the number of component types. $\alpha_c$ is the weight coefficient for $L_\text{dnc}^c$, which is determined by the number of samples on the corresponding component type. 

In addition to SRJ and DNC, we retain image-text contrastive learning (ITC) \cite{ref15,ref16} that used in training the open-source pretrained model as a fundamental pretraining task to avoid the corruption of pretrained representations. Therefore, the total pretraining loss of VLP-TL is:
\begin{equation} \label{totalloss}
L_\text{vlp-tl} = \lambda_{1}L_\text{itc} + \lambda_{2}L_\text{srj} + \lambda_{3}L_\text{dnc} \,.
\end{equation}
where $L_\text{itc}$ is the loss of ITC. $\lambda_{1},\lambda_{2},\lambda_{3}$ are the weight coefficients for the losses of the three pertaining tasks.

\subsection{Transition Through PTS}
When transferring the pretrained image encoder to defect detection, two latent factors lead to suboptimal performance. 1) Data discrepancy: From the perspective of training data, pretraining uses instance-level images where the object is mainly in the center of the image and occupies most of the area, while defect detection uses original UAV patrol images where the object is randomly positioned and occupies a small area. 2) Objective inconsistency: Three pretraining tasks, namely SRJ, DNC, and ITC, are used in pretraining to optimize the image encoder, while object classification and bounding box regression are used to optimize the detector in defect detection. To mitigate the gap in training data and training objectives between pertaining and defect detection, PTS introduces an additional transition stage to progressively transfer the pretrained image encoder to defect detection, as shown in Stage 2 of Fig.~\ref{fig_1}.

\begin{figure}[!t]
\centering
\includegraphics[width=3.4in]{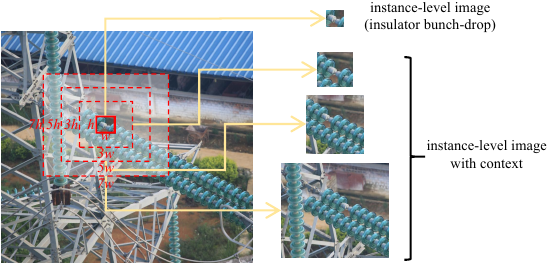}
\caption{The method to obtain instance-level images with context.}
\label{fig_5}
\end{figure}

From the perspective of training data, PTS introduces instance-level images with context for a smoother transition. An instance-level image with context is an image containing both an object of interest and its surrounding environment. For example, the instance-level image with context about the insulator bunch-drop in Fig.~\ref{fig_5} not only contains the vacant position of an insulator but also includes adjacent insulator strings around it. To obtain instance-level images with context we perform multiscale cropping on the original UAV patrol images from the TLDD dataset, as shown in Fig.~\ref{fig_5}. Given a defect region of size $h\times w$ in the UAV patrol image, we define three regions centered on the defect with sizes $3h\times 3w$, $5h\times 5w$, and $7h\times 7w$. Within the three regions, we randomly crop three instance-level images with context whose sizes are $h_1\times w_1$, $h_2\times w_2$, and $h_3\times w_3$. Specifically, $h \leq h_1 \leq 3h$, $3h \leq h_2 \leq 5h$, $5h \leq h_3 \leq 7h$, and $w \leq w_1 \leq 3w$, $3w \leq w_2 \leq 5w$, $5w \leq w_3 \leq 7w$. During the transition stage, we mix the obtained instance-level images with context into the multimodal dataset used in pretraining for training. Since the instance-level images with context better match the style of original UAV patrol images, they can bridge the gap in training data between pretraining and defect detection.

From the perspective of training objectives, PTS retains the ITC pretraining task in the transition stage. The ITC is reused to preserve pretrained representations and alignment between the two modalities, while SRJ and DNC are continually implemented to capture more fine-grained visual cues from instance-level images with context. Thus, the total loss during the transition stage maintains the same formulation as in the pretraining stage (\emph{i.e.}, Eq.\ref{totalloss}). 

Through bridging pretraining and defect detection through coordinated data and objective adaptation, PTS can effectively improve defect detection accuracy without introducing additional parameters or compromising inference speed.

\subsection{Transfer to Defect Detection}
After pretraining and transition, we transfer the obtained image encoder to defect detection, as shown in Stage 3 of Fig.~\ref{fig_1}.

We construct a specialized detection decoder following the architecture of ViTDet \cite{ref41} on the top of the pretrained image encoder. Specifically, we first use the image encoder as the backbone of the defect detector and add a simple feature pyramid network (FPN) on the output end of it to provide multiscale features for defect detection. Then we build a region proposal network (RPN) and a box head to process the multiscale features to locate and classify defects that appear in the input images. Finally, the transfer to defect detection is complete after the entire detector is trained on the annotated original UAV patrol images from the TLDD dataset.

\section{Experiments and Analysis} \label{sec4}
In this section, we first introduce the datasets used in our experiments and the experimental settings. Then, we evaluate the proposed method by comparing it with other existing methods and implementing ablation experiments.

\subsection{Datasets}
The experimental images used in this article are collected by UAVs equipped with high-definition image transmission systems during patrol inspections. We manually filter and remove the blurred images to ensure data quality. Afterward, we select a part of defect-containing images and annotate the defects within them to build the TLDD dataset for training and testing the detectors. For the remaining images, we annotate and crop both normal and defect components and external interference defects to build the multimodal dataset for pertaining.

The multimodal dataset is a pertaining dataset curated for the proposed VLP-TL algorithm. It is designed to pretrain the image encoder with image-text pairs containing rich multimodal information. The multimodal dataset contains 23,391 samples in the form of image-text pairs and covers 24 common normal and defect categories about adjusting boards, composite insulators, disc suspension insulators, glass insulators, grading rings, yoke plates, suspension clamps, shielded rings, shockproof hammers, weight, bird nests, foreign bodies, etc. It's noteworthy that we also reserve the category annotation for each image-text pair so that we can perform SRJ and DNC pretraining tasks efficiently. Some typical image-text pairs and the corresponding category annotations in the multimodal dataset are shown in Table \ref{tab1}. 

\begin{table}[!t]
\caption{Some Samples in the Multimodal Dataset for Pretraining}
\centering
\label{tab1}
\begin{tabular}{m{2.4cm}<{\centering}m{2.6cm}<{\centering}m{2.5cm}<{\centering}m{2.4cm}<{\centering}}
\toprule
Instance-level Images & Alt-texts & Category Annotations \\ \midrule
\includegraphics[width=0.8in]{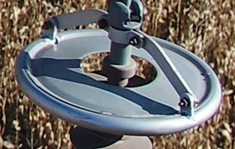}
 & There is a normal grading ring in the image. & normal grading ring\\ \midrule
\includegraphics[width=0.8in]{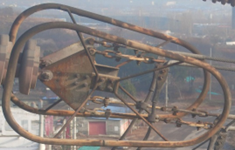}
 & A rusted shielded ring on the aerial inspection image. & shielded ring corrosion\\ \midrule
\includegraphics[width=0.8in]{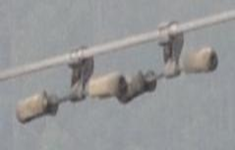}
 & Two shockproof hammers are closely adjacent to each other. & shockproof hammer intersection\\
\bottomrule
\end{tabular}
\end{table}

\begin{figure}[!t]
\centering
\includegraphics[width=3.4in]{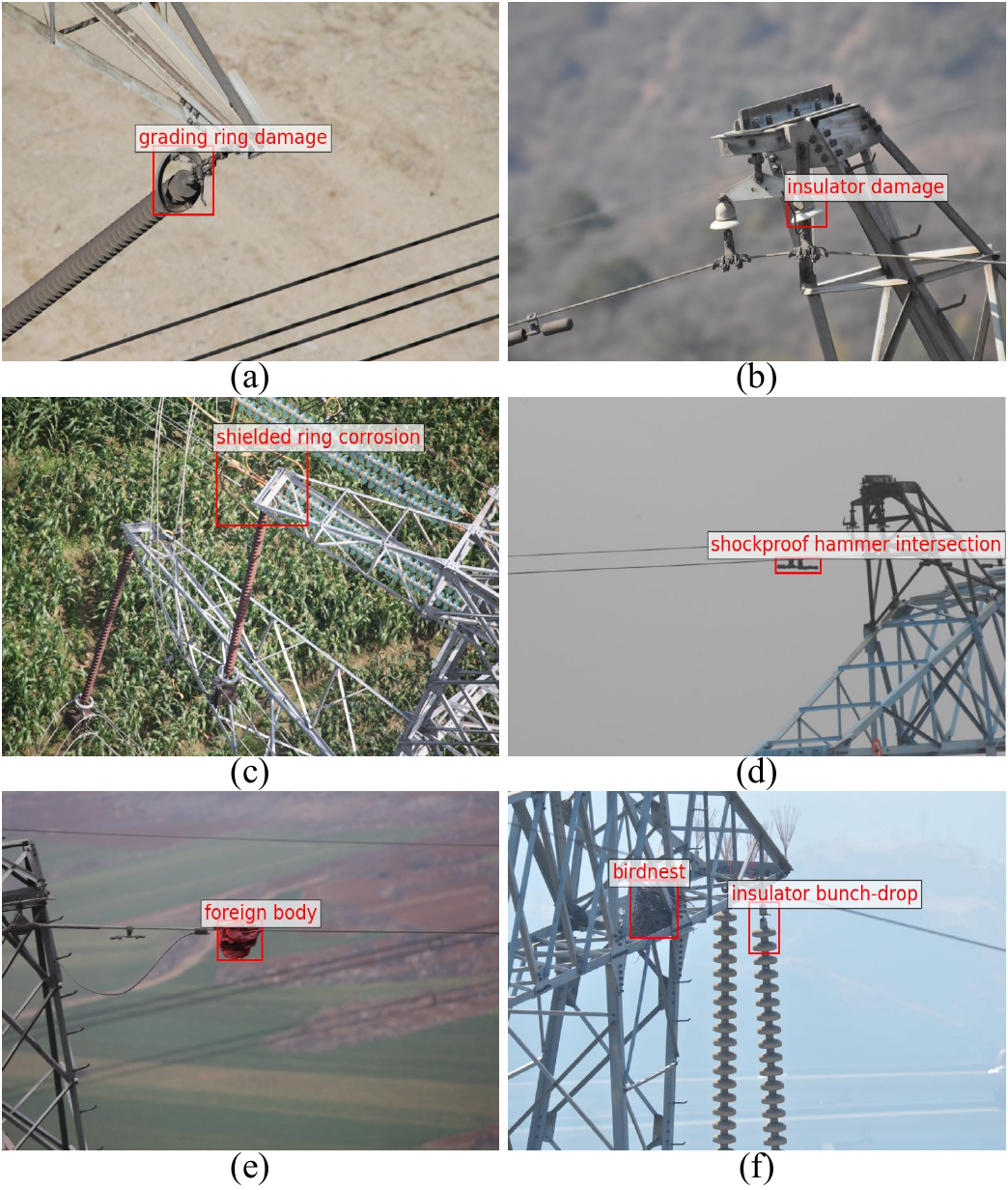}
\caption{Data examples in the TLDD dataset. (a) A defect of grading ring damage. (b) A defect of insulator damage. (c) A defect of shielded ring corrosion. (d) A defect of shockproof hammer intersection. (e) A defect of foreign body. (e) A defect of bird nest and a defect of insulator bunch-drop.}
\label{fig_6}
\end{figure}

The TLDD dataset is a specialized dataset for transmission line defect detection. We use the TLDD dataset to train different types of detectors and evaluate their performance on defect detection. In our method, the TLDD dataset is used to provide instance-level images with context for transition and transfer the pretrained image encoder to the defect detection task. The TLDD dataset includes a total of 1,830 images with 7 defect categories: 1) bird nest (BN); 2) foreign body (FB); 3) grading ring damage (GRD); 4) insulator bunch-drop (IBD); 5) insulator damage (ID); 6) shielded ring corrosion (SRC); 7) shockproof hammer intersection (SHI). The training set and testing set are divided with a 9:1 ratio. Some samples in the TLDD dataset and their annotations are shown in Fig.~\ref{fig_6}.

\subsection{Experimental Settings}\label{dnc_subtasks}
We implement the experiments using the PyTorch framework on a workstation with two NVIDIA RTX3090 GPUs. Before pretraining, we use the open-source pretrained weights of CN-CLIP ViT-B-16 \cite{ref39} to initialize the image encoder and text encoder. In the stage of pretraining, we resize the resolution of each input image to $224\times224$. We pretrain both encoders for 30 epochs with the Adam optimizer. The initial learning rate is set to $1\times10^{-6}$. The batch size is adjusted to 120 according to the available memory of the used devices. In the stage of transition, we continue to optimize the pretrained image encoder and text encoder for 15 epochs while keeping other hyperparameters unchanged. In the stage of defect detection, we adopt the multiscale training strategy to adapt the image encoder to defect detection, requiring higher input resolution. We train the defect detector for 50 epochs with the AdamW optimizer. The initial learning rate is $1\times10^{-5}$ for the backbone and $1\times10^{-4}$ for other modules in the detection decoder.

To evaluate the detection accuracy, we adopt mean average precision (mAP) as the main metric. mAP is the average value of average precision (AP) across all categories, and the calculation of AP and mAP can be formulated as:
\begin{equation}
P=\frac{T_\mathrm{p}}{T_\mathrm{p}+F_\mathrm{p}} 
\end{equation}
\begin{equation}
R=\frac{T_\mathrm{p}}{T_\mathrm{p}+F_\mathrm{n}} 
\end{equation}
\begin{equation}
AP=\int_0^1 P(R)\mathrm{d}R 
\end{equation}
\begin{equation}
mAP=\frac{1}{m}\sum_{i=1}^{m}AP_{i}
\end{equation}
where $T_\mathrm{p}$ is the number of correctly detected defects, $F_\mathrm{p}$ is the number of wrongly detected defects, $F_\mathrm{n}$ is the number of undetected defects, $P$ is the precision of the category, $R$ is the recall of the category, and $AP_i$ is the AP of the $i$-th category. $m$ is the number of defect categories and $m=7$ in the TLDD dataset. According to different intersection over union (IoU) thresholds used in calculating AP, mAP can be divided into $\text{mAP}_{50}$, $\text{mAP}_{75}$, and $\text{mAP}_{50:95}$. Among them, $\text{mAP}_{50}$ and $\text{mAP}_{75}$ represents the mAP values calculated under IoU thresholds of 0.5 and 0.75, respectively. $\text{mAP}_{50:95}$ represents the average value of all mAP values calculated under IoU thresholds increasing from 0.5 to 0.95 with a step size of 0.05. Additionally, we use the number of parameters (Params) to evaluate the detector size and frame per second (FPS) to evaluate the inference speed of the detector.

\begin{table*}[!t]
  \caption{Detection Results on the TLDD Dataset}
  \label{tab2}
  \begin{tabular*}{\hsize}{@{\extracolsep{\fill}}ccccccc}
    \toprule
    Detectors & Backbone & Params(M) & FPS & $\text{mAP}_{50}$(\%) & $\text{mAP}_{75}$(\%) & $\text{mAP}_{50:95}$(\%)\\
    \midrule
    \emph{DETR-like Detectors} & & & & & & \\
    DETR & ResNet-50 & 41.3 & 18.2 & 70.9 & 50.2 & 44.1\\
    Deformable DETR & ResNet-50 & 39.8 & 14.7 & 75.6 & 44.6 & 44.1\\
    Conditional DETR & ResNet-50 & 43.7 & 15.4 & 72.5 & 30.6 & 33.6\\
    PA-DETR & ResNet-50 & 41.6 & 18.0 & 76.4 & 50.8 & 46.4\\
    \midrule
    \emph{Single-stage Detectors} & & & & & \\
    RetinaNet & ResNet-50 & 36.5 & 37.4 & 68.8 & 40.9 & 39.3\\
    RTMDet & CSP-NeXt & 52.3 & 25.1 & 70.0 & 50.2 & 44.9\\
    YOLOv8 & CSP-DarkNet & 43.6 & 56.4 & 65.7 & 45.4 & 41.8\\
    improved YOLOv7 & CSP-DarkNet & 48.3 & 41.2 & 71.4 & 50.6 & 44.7\\  
    \midrule
    \emph{Two-stage Detectors} & & & & & \\
    Faster RCNN & ResNet-50 & 41.4 & 15.5 & 65.3 & 44.9 & 42.8\\
    Cascade RCNN & ResNet-50 & 69.2 & 15.2 & 66.7 & 45.6 & 43.6\\
    Sparse RCNN & ResNet-50 & 106.3 & 16.4 & 73.3 & 48.9 & 44.7\\
    ViTDet & ViT-B & 116.0 & 12.4 & 71.6 & 52.3 & 44.9\\
    Ours & ViT-B & 112.5 & 12.2 & 78.1 & 54.9 & 47.2\\
  \bottomrule
\end{tabular*}
\end{table*}

\begin{table*}[!t]
\caption{Ablation Analysis of pretraining Tasks}
\centering
\label{tab3}
\begin{tabularx}{\linewidth}{>{\centering\arraybackslash}X>{\centering\arraybackslash}X>{\centering\arraybackslash}X>{\centering\arraybackslash}X>{\centering\arraybackslash}X>{\centering\arraybackslash}X>{\centering\arraybackslash}X>{\centering\arraybackslash}X>{\centering\arraybackslash}X>{\centering\arraybackslash}X>{\centering\arraybackslash}X>{\centering\arraybackslash}X}
\toprule
\multirow{2}{*}{VLP-TL} & \multirow{2}{*}{PTS} & \multirow{2}{*}{Params(M)} & \multirow{2}{*}{$\text{mAP}_{50}$(\%)} & \multicolumn{7}{c}{AP(\%)} \\  
\cmidrule{5-11} & & & & BN & FB & GRD & IBD & ID & SRC & SHI \\
\midrule
 - & - & 112.5 & 74.2 & 84.7 & 64.8 & 53.9 & 83.0 & 51.1 & 86.2 & 95.5       \\
 $\surd$ & - & 112.5 & 76.5 & 87.3 & 68.9 & 57.5 & 84.0 & 52.9 & 87.6 & 97.2 \\ 
 - & $\surd$ & 112.5 & 75.4 & 86.0 & 66.3 & 55.8 & 84.2 & 52.5 & 87.0 & 96.1 \\
 $\surd$ & $\surd$ & 112.5 & 78.1 & 89.8 & 71.7 & 60.4 & 84.9 & 54.2 & 88.1 & 97.6 \\
\bottomrule
\end{tabularx}
\end{table*}

\subsection{Experiments on the TLDD Dataset}
To verify the effectiveness of the defect detector built in Section \ref{sec3}, we use the TLDD dataset to compare its performance with other detectors commonly used in the domain of electric power computer vision, and the results are shown in Table~\ref{tab2}. Specifically, Faster RCNN \cite{ref40}, Cascade RCNN \cite{ref55}, Sparse RCNN \cite{ref56}, and ViTDet \cite{ref41} are two-stage detectors that first generate candidate boxes with RPN and refine them with a subsequent neural network for classification and location. RetinaNet \cite{ref57}, RTMDet \cite{ref58}, and YOLOv8 \cite{ref59} are single-stage detectors that directly use the features extracted by the backbone for classification and location without generating candidate boxes. DETR \cite{ref60}, Deformable DETR \cite{ref61}, and Conditional DETR \cite{ref62} are DETR-like detectors that are a family of end-to-end detectors regarding object detection as a collection prediction problem. In addition, the improved YOLOv7 \cite{ref54} and PA-DETR \cite{ref24} are the latest proposed single-stage and DETR-like detectors for transmission line defect detection. Although our detector has more parameters and lower FPS compared with other detectors of the three types, it achieves significantly higher detection accuracy, proving the advantage of jointly utilizing multimodal information to train defect detectors. In particular, compared with the ViTDet with almost the same backbone architecture pretrained on the ImageNet \cite{ref43}, the $\text{mAP}_{50}$, $\text{mAP}_{75}$, and $\text{mAP}_{50:95}$ of our detector are improved by 6.5\%, 2.6\%, and 2.3\%, respectively. This result demonstrates the effectiveness of incorporating knowledge from multimodal data into the detector backbone via VLP-TL and PTS.

\begin{figure*}[!t]
\centering
\includegraphics[width=6.4in]{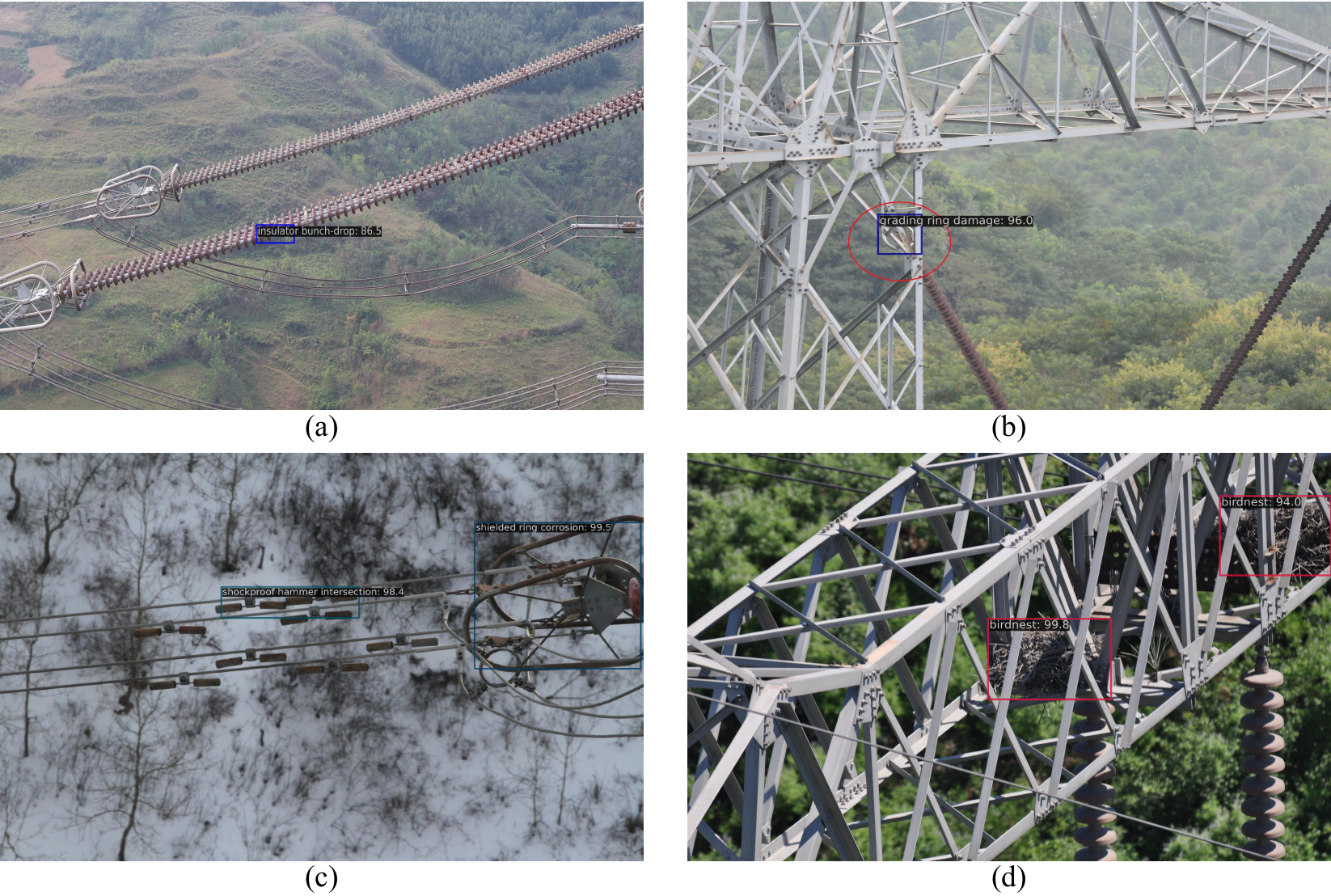}
\caption{Detection results of the proposed detector. (a) A small-sized insulator bunch-drop defect. (b) A damaged grading ring occluded by the tower. (c) A shockproof hammer intersection and a shielded ring corrosion defect in the complex background. (d) Two bird nests in the complex background.}
\label{fig_7}
\end{figure*}

Fig.~\ref{fig_7} visualizes the detection results of our detector on the TLDD dataset. It shows that our detector accurately localizes the insulator bunch-drop defect even if it only occupies a small area in the image in Fig.~\ref{fig_7}(a). In Fig.~\ref{fig_7}(b), our detector successfully finds the damaged grading ring partially occluded by the tower. In Fig.~\ref{fig_7}(c) and Fig.~\ref{fig_7}(d), our detector also locates the shockproof hammer intersection and bird nests from the interferential background. In general, our detector can effectively overcome the problem of insufficient defect-related visual information in UAV patrol images by jointly utilizing information from the two modalities, thus detecting defects more accurately.

\subsection{Ablation Analysis}
In order to verify the effectiveness of the proposed VLP-TL and PTS, we design a set of ablation experiments for comparison. In particular, we further analyze some important factors involved in VLP-TL and PTS.

\emph{1) Ablation Study:} The results of the ablation experiment are shown in Table~\ref{tab3}. When neither VLP-TL nor PTS is utilized, the baseline detector is a ViTDet whose backbone is initialized with the pretrained weights of CN-CLIP. Based on the baseline detector, we introduce the VLP-TL and PTS separately to investigate the effect of each improvement scheme.

When VLP-TL is employed to pretrain the backbone of the detector, $\text{mAP}_{50}$ of the detector is improved by 2.3\% while the corresponding APs of all categories are improved to varying degrees. This verifies that VLP-TL can effectively mine the domain knowledge contained in the multimodal dataset, thereby making the backbone more suitable for the transmission line scenario. On the other hand, when only PTS is used to transfer the pretrained image encoder to defect detection, $\text{mAP}_{50}$ is improved by 1.2\%. But overall, when only using PTS, the increases in $\text{mAP}_{50}$ and AP of each category are not as good as those of VLP-TL. This is because there is a lack of sufficient exposure to instance-level images in the pretraining stage. Specifically, directly introducing instance-level images with context increases the difficulty of aligning the two modalities, thereby hindering the image encoder from learning professional knowledge. Additionally, it can be seen that neither improvement scheme introduces additional parameters, because they are improved from the perspective of training strategy without modifying the original detector architecture. Our method is improved by combining the two improvement schemes, resulting in a 3.9\% increase in $\text{mAP}_{50}$ and significant increases in AP across all categories.

\emph{2) Analysis of VLP-TL:} A series of experiments are performed from the perspective of pretraining data and pretraining tasks to validate the role of the multimodal dataset and two novel pretraining tasks in defect detection. They are elaborated as follows:

\begin{table}[!t]
\caption{Ablation Analysis of pretraining Data}
\centering
\label{tab4}
\begin{tabularx}{\linewidth}{>{\centering\arraybackslash}X>{\centering\arraybackslash}X>{\centering\arraybackslash}X>{\centering\arraybackslash}X>{\centering\arraybackslash}X}
\toprule
 \multicolumn{2}{c}{Pretraining Data} & \multirow{2}{*}{$\text{mAP}_{50}$} & \multirow{2}{*}{$\text{mAP}_{75}$} & \multirow{2}{*}{$\text{mAP}_{50:95}$}\\ 
 \cmidrule{1-2} image & alt-text & & & \\
\midrule
 $\surd$ & - & 71.9 & 46.4 & 40.9  \\
 $\surd$ & $\surd$ & 75.4 & 48.1 & 42.9  \\ 
\bottomrule
\end{tabularx}
\end{table}

\textbf{Analysis of pretraining data:} To evaluate the impact of multimodal information in the pretraining data on defect detection, we compare the defect detection performance of VLP that jointly uses vision and language modalities with traditional supervised pretraining using only the vision modality. Concretely, we first build a classification dataset with all images and category annotations from the multimodal dataset. Then, we use the multimodal dataset and the classification dataset to pretraining two image encoders with image classification and ITC, respectively. Finally, we build two defect detectors based on the image encoders and test their detection accuracy, as shown in Table~\ref{tab4}. It can be found that when we introduce the language modality beyond the vision modality, the $\text{mAP}_{50}$, $\text{mAP}_{75}$, and $\text{mAP}_{50:95}$ are improved by 3.5\%, 1.7\%, and 2.0\%, respectively. It proves that the detector benefits from the additional information introduced by the language modality, so the detection accuracy is effectively improved.

\begin{table}[!t]
\caption{Ablation Analysis of pretraining Tasks}
\centering
\label{tab5}
\begin{tabularx}{\linewidth}{>{\centering\arraybackslash}X>{\centering\arraybackslash}X>{\centering\arraybackslash}X>{\centering\arraybackslash}X>{\centering\arraybackslash}X>{\centering\arraybackslash}X}
\toprule
 ITC & SRJ & DNC & $\text{mAP}_{50}$ & $\text{mAP}_{75}$ & $\text{mAP}_{50:95}$ \\  
\midrule
 - & - & - & 74.2 & 44.9 & 42.3  \\
 $\surd$ & - & - & 75.4 & 48.1 & 42.9  \\ 
 - & $\surd$ & - & 41.8 & 27.3 & 23.9   \\
 - & - & $\surd$ & 41.8 & 22.4 & 21.7   \\
 $\surd$ & $\surd$ & - & 75.9 & 48.7 & 43.1 \\
 $\surd$ & - & $\surd$ & 76.0 & 49.5 & 44.6 \\
 - & $\surd$ & $\surd$ & 49.0 & 28.5 & 27.2 \\
 $\surd$ & $\surd$ & $\surd$ & 76.5 & 50.4 & 45.9 \\
\bottomrule
\end{tabularx}
\end{table}

\begin{figure}[!t]
\centering
\includegraphics[width=2.8in]{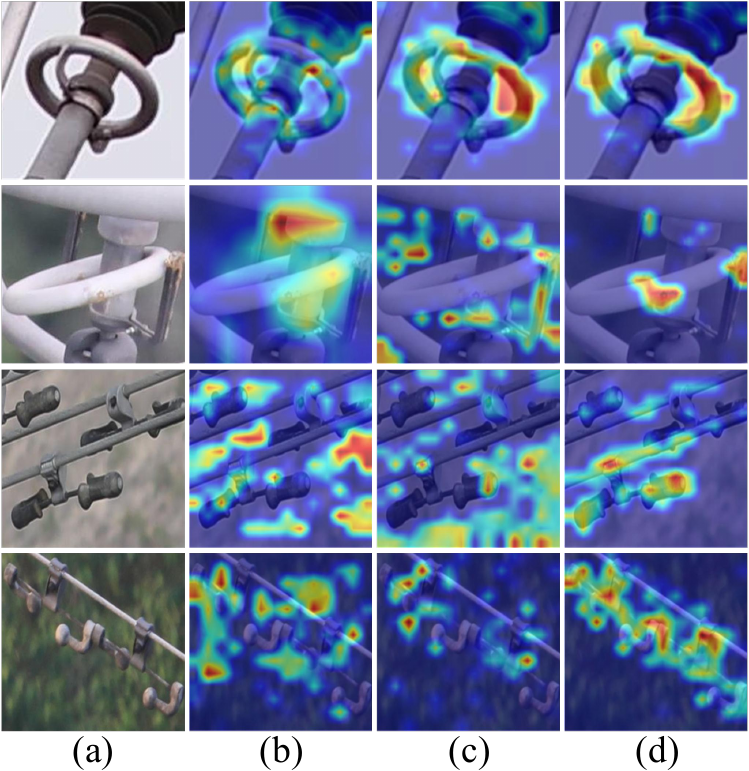}
\caption{Some images from the multimodal dataset and corresponding attention maps generated by different image encoders. (a) Original images. (b) Attention maps of the ViT pretrained on ImageNet. (c) Attention maps of the original ViT in CN-CLIP. (d) Attention maps of the ViT in CN-CLIP pretrained with VLP-TL.}
\label{fig_8}
\end{figure}

\textbf{Analysis of pretraining tasks:} To examine the effect of various pretraining tasks on defect detection, we performed an ablation analysis of the pretraining tasks within the VLP-TL algorithm. The results are shown in Table~\ref{tab5}. It is easy to find that when ITC is not used, the detection accuracy is significantly lower than the case without pretraining, whether using SRJ or DNC. This demonstrates that if ITC is not used as the fundamental pretraining task, the original pretraining representations will be corrupted by other pretraining tasks, which will have an adverse impact on defect detection. On the contrary, when DNC or SRJ tasks are introduced on the basis of ITC, the $\text{mAP}_{50}$, $\text{mAP}_{75}$, and $\text{mAP}_{50:95}$ can be improved to varying degrees. This verifies the effectiveness of the two proposed pretraining tasks. In other words, both SRJ and DNC can further explore the power-related professional knowledge contained in the multimodal pretraining data based on the alignment of transmission line images and alt-texts through ITC. Furthermore, the combination of three pretraining tasks result in 2.3\%, 5.5\%, and 3.6\% increase in $\text{mAP}_{50}$, $\text{mAP}_{75}$, and $\text{mAP}_{50:95}$, which verifying the effectiveness of the VLP-TL algorithm. 

To intuitively observe the effect of pretraining the image encoder using the proposed VLP-TL algorithm, we select some typical images from the multimodal dataset and visualize the attention maps generated by different image encoders in Fig.~\ref{fig_8}. Specifically, we compare the attention maps generated by three types of ViT image encoders: ViT pretrained on ImageNet, original ViT in CN-CLIP, and ViT in CN-CLIP pretrained with VLP-TL. The first and second rows of Fig.~\ref{fig_8} are normal and damaged grading rings, respectively. As can be seen from the first row, the image encoder pretrained with VLP-TL can more accurately focus on the ring-shaped area corresponding to the grading ring. More impressively, the image encoder pretrained with VLP-TL accurately locates the damaged grading ring and two breakpoints in the second row, while two other image encoders fail. The third and fourth rows of Fig.~\ref{fig_8} show normal shockproof hammers and a shockproof hammer intersection defect, respectively. In the third row, the image encoder pretrained with VLP-TL can find the shockproof hammer while also noting the two different cables, thereby avoiding the misjudgment of the intersection defect. In the fourth row, the image encoder pretrained with VLP-TL correctly finds the two shockproof hammers and their intersection point. However, the two other image encoders both fail to locate the shockproof hammers in the third and fourth rows. In summary, through pretraining with VLP-TL, the image encoder is more sensitive to components and defects in transmission lines, indicating the benefit of multimodal information for the image encoder.

\begin{table}[!t]
\caption{Ablation Analysis of pretraining Data}
\centering
\label{tab6}
\begin{tabularx}{\linewidth}{>{\centering\arraybackslash}X>{\centering\arraybackslash}X>{\centering\arraybackslash}X>{\centering\arraybackslash}X}
\toprule
 \multirow{2}{*}{\makecell{Number of\\used sizes}} & \multirow{2}{*}{$\text{mAP}_{50}$} & \multirow{2}{*}{$\text{mAP}_{75}$} & \multirow{2}{*}{$\text{mAP}_{50:95}$}\\ 
 & & & \\
\midrule
 1 & 76.7 & 51.7 & 46.3  \\
 2 & 77.3 & 53.5 & 46.8  \\ 
 3 & 78.1 & 54.9 & 47.2  \\
 4 & 77.6 & 54.1 & 46.5  \\
 5 & 76.5 & 52.6 & 45.7  \\
\bottomrule
\end{tabularx}
\end{table}

\emph{3) Analysis of PTS:} Since the training objective selection in the transition stage shows a similar pattern in the subsequent defect detection as in Table~\ref{tab5}, we continue to use ITC, SRJ, and DNC as the training objectives in this stage and do not elaborate on them again. In the following, we focus on the size selection of instance-level images with context in the proposed PTS. 

We considered five candidate sizes for generating instance-level images with context: $h_1\times w_1$, $h_2\times w_2$, $h_3\times w_3$, $h_4\times w_4$, and $h_5\times w_5$. Specifically, $h \leq h_1 \leq 3h$, $3h \leq h_2 \leq 5h$, $5h \leq h_3 \leq 7h$, $7h \leq h_4 \leq 9h$, $9h \leq h_5 \leq 11h$, and $w \leq w_1 \leq 3w$, $3w \leq w_2 \leq 5w$, $5w \leq w_3 \leq 7w$, $7w \leq w_4 \leq 9w$, $9w \leq w_5 \leq 11w$, where $h$ and $w$ are the original height and width of a defect region. We add instance-level images with context to the multimodal dataset to build five datasets in order of size from small to large, that is, from mixing in images of only one size ($h_1\times w_1$) to mixing in images of five sizes (from $h_1\times w_1$ to $h_5\times w_5$) simultaneously. We then use the five datasets during the transition stage for training separately and measure the corresponding defect detection performance of the image encoders after transition, the results are shown in Table~\ref{tab6}. It demonstrates that detection accuracy is gradually improved as the used sizes increase from one to three; and when the number of used sizes reaches three, the corresponding detection accuracy reaches a peak and then begins to decline. The reason for this phenomenon is that when the number of used sizes is less than three, the defect-related information in the instance-level image with context is appropriately preserved to facilitate alignment with the alt-text and finding more helpful visual cues. However, when the number of used sizes is greater than three, the newly introduced instance-level images with context contain too many irrelevant objects or exceed the size of the original UAV patrol image in the TLDD dataset, increasing the difficulty of modality alignment and mining useful information. Therefore, we use three sizes of instance-level images with context for transition.

\section{Conclusion} \label{sec5}
This article proposes a VLP-based transmission line defect detection method to mitigate the problem of insufficient defect-related visual information in UAV patrol images. In the proposed method, we implement two additional stages before defect detection to learn from multimodal data and transfer the acquired knowledge. First, in the pretraining stage, we pretrain an image encoder via VLP-TL, a VLP algorithm designed for the transmission line scenario. Specifically, VLP-TL contains two customized pretraining tasks besides ITC, namely SRJ and DNC, to explore professional knowledge and visual cues from a large number of transmission line image-text pairs. Then, we propose PTS to endow the pretrained image encoder with better transfer performance in defect detection by implementing the transition stage. In the transition stage, we bridge the gap between pretraining and defect detection by introducing images with styles closer to the original UAV patrol images used for defect detection and retaining the tasks used in pretraining as training objectives. After the pretraining and transition stage, we use the obtained image encoder as the backbone to build a defect detector that can accurately detect defects under the conditions of small size, partial occlusion, and complex background.

The presented work provides new ideas for the subsequent development of VLP-based defect detectors specific to UAV patrol inspection of transmission lines. We will further exploit distinctive relations and characteristics hidden in the UAV patrol images to customize pretraining tasks for the transmission line scenario. Moreover, we hope to further expand our research to other defect detection tasks in the generation, transformation, and distribution of power systems, and develop more general VLP algorithm and defect detector for the domain of electric power.

\bibliographystyle{IEEEtran}
\bibliography{bibtex}


\end{document}